\documentclass[journal]{IEEEtran}
\ifCLASSINFOpdf
\else
\fi
\usepackage{lineno,hyperref}
\usepackage{graphicx}
\usepackage{multirow}
\usepackage{hhline}
\usepackage{chngpage}
\usepackage{longtable}
\usepackage{tabularx}
\usepackage{csquotes}
\usepackage{textcomp}

\usepackage[monochrome]{color}

\newcolumntype{R}[1]{>{\raggedleft\arraybackslash}p{#1}}

\usepackage[caption=false,font=footnotesize]{subfig}

\begin{document}

%
\title{Towards Storytelling from \\Visual Lifelogging: An Overview}
%
%
%

\author{Marc~Bola\~nos$^*$,~
        Mariella~Dimiccoli$^*$,~
        and~Petia~Radeva
\thanks{M. Bola\~nos, marc.bolanos@ub.edu, M. Dimiccoli, mariella.dimiccoli@cvc.uab.es, P. Radeva, petia.ivanova@ub.edu - Universitat de Barcelona, Barcelona, Spain and Computer Vision Center, Bellaterra, Spain. $^*$The first two authors contributed equally to this work.}
}

%
%

\markboth{Journal of Transactions on Human-Machine Systems July~2015}%
{Shell \MakeLowercase{\textit{et al.}}: Bare Demo of IEEEtran.cls for Journals}
%



\maketitle

\begin{abstract}
Visual lifelogging consists of acquiring images that capture the daily experiences of the user by wearing a camera over a long period of time. The pictures taken offer considerable potential for knowledge mining concerning how people live their lives, hence, they open up new opportunities for many potential applications in fields including healthcare, security, leisure and the quantified self. However, automatically building a story from a huge collection of unstructured egocentric data presents major challenges.  
This paper provides a thorough review of advances made so far in egocentric data analysis, and in view of the current state of the art, indicates new lines of research to move us towards storytelling from visual lifelogging.

\end{abstract}


%
\IEEEpeerreviewmaketitle

\vspace{-1.5em}
\section{Introduction}

\begin{figure}[!t]
\begin{center}
\includegraphics[width=0.9\columnwidth]{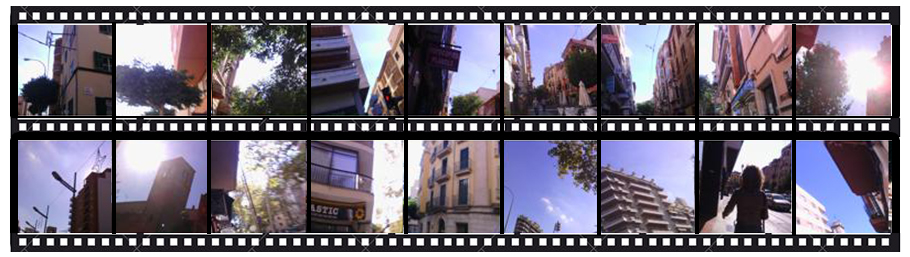}
\caption{Example of a sequence acquired by the Narrative Clip wearable camera while the user is walking down a street. The temporal leaps between neighbouring pictures produced by photographic cameras are common in dynamic environments and make the extraction of information from closely spaced images very difficult.
}
\label{fig:low_resolution}
\end{center}
\vspace{-1em}
\end{figure}

\IEEEPARstart{L}{ifelogging} consists of a user continuously recording their everyday experiences, typically via wearable sensors including accelerometers and cameras, among others. When the visual signal is the only one recorded, typically by a wearable camera, it is referred to as \textit{visual lifelogging}. This is a trend that is rapidly increasing thanks to advances in wearable technologies over recent years. Nowadays, wearable cameras are very small devices that can be worn all-day long and automatically record the everyday activities of the wearer in a passive fashion, from a first-person point of view. As an example, Fig. \ref{fig:low_resolution} shows pictures taken by a person walking down a street while wearing such a camera.

\begin{figure}
\centering
  \begin{tabular}{cccc}
 \includegraphics[trim=0.2cm 0 0.2cm 0, clip=true, width=2.2cm,height=1.6cm]{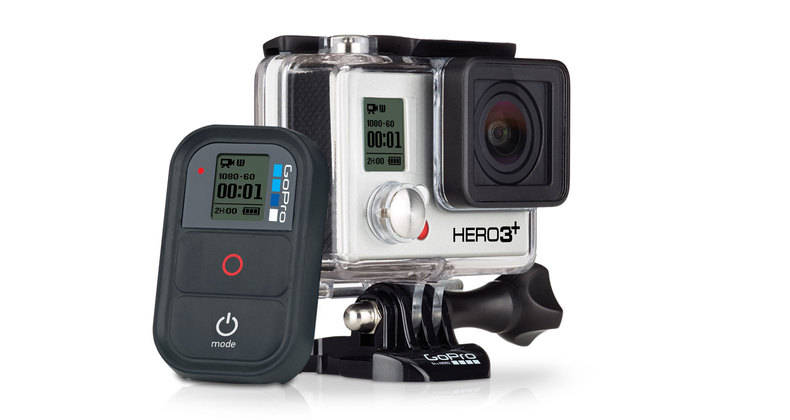}&\hspace{-1em}
 \includegraphics[trim=0.8cm 0 0.8cm 0, clip=true, width=1.6cm,height=2cm]{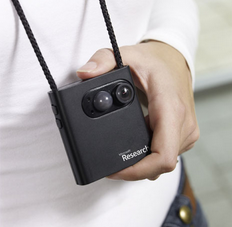}&
 \includegraphics[trim=0 0 0 0, clip=true, width=2cm,height=2cm]{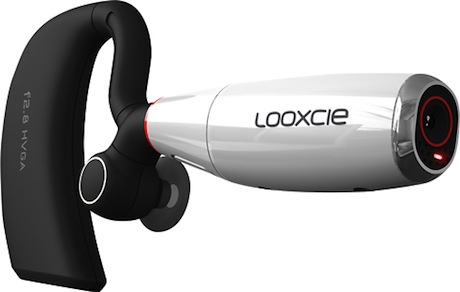}&
 \includegraphics[width=1.8cm,height=2cm]{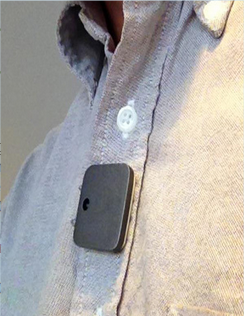}\\
 (a)&(b)&(c)&(d)
    \end{tabular}
    \caption{Examples of wearable cameras on the market: (a) GoPro (2002). (b) SenseCam (2005). (c) Looxcie (2011). (d) Narrative Clip (2013).}
	\label{fig:wearable_cameras}
    \vspace{-1em}
\end{figure}

Most wearable cameras on the market like  GoPro, MeCam, Looxcie or Google Glass (see Fig. \ref{fig:wearable_cameras} (a) and (c)) are video cameras, which have relatively High Temporal Resolution (HTR) (e.g. from 25 up to 60 frames per second) and are more suitable to record specific moments, such as cooking or doing sports. 
A limited number of wearable cameras, such as Narrative Clip and SenseCam (see Fig. \ref{fig:wearable_cameras} (b) and (d)) are photographic cameras, which have Low Temporal Resolution (LTR) (2-3 frames per minute), and hence are more suitable for acquiring data over long periods of time.
On the one hand, data recorded at specific moments with video cameras offer potential for in-depth analysis of daily or special activities, allowing to capture even \textit{how} something happened. On the other hand, data acquired over long periods of time, commonly called \textit{visual lifelogs},  offer considerable potential for inferring knowledge about e.g. behaviour patterns, 
and hence enable many applications that would not be possible with HTR cameras. As shown by Doherty et al. \cite{Doherty2013Wearable}, visual lifelogs captured through a SenseCam, which as opposed to video cameras can capture the whole day, could be used to prevent non-communicable diseases associated with unhealthy trends and risky profiles (such as obesity or depression, among others). Additionally, they could also help prevent cognitive and functional decline in elderly people \cite{doherty2012experiences,hodges2006sensecam,Lee08lifeloggingmemory}. 
However, visual lifelogs present a significant challenge for automatic visual analysis. Indeed, due to the free motion of the camera and to its LTR, abrupt changes in lighting conditions and image content are very frequent (see Fig. \ref{fig:low_resolution}). In such situations, computer vision techniques based on temporal coherence and motion estimation become unreliable. Recognition algorithms have to cope with the huge variety of objects that appear. In addition, due to the non-intentional nature of the pictures captured, they generally contain severely occluded objects, artefacts such as blurring or light saturation \cite{tan2014understanding} and a large number of non-informative images that capture non-meaningful information such as walls, the sky, parts of objects, etc. Furthermore, the sheer number of data that a visual lifelog consists of and the rate at which they increase (up to 2,000 images per day or around 800,000 images every year) imposes a need for efficient methods to extract and locate
relevant content concerning the wearer from the photo stream. Regarding HTR cameras, if they were employed for a lifelog analysis, the problem of the amount of data would be even more acute, and would additionally imply the need of huge computational resources.

\begin{figure}[b!]
 	\vspace{-2em}
	\begin{center}

	\includegraphics[width=0.8\columnwidth]{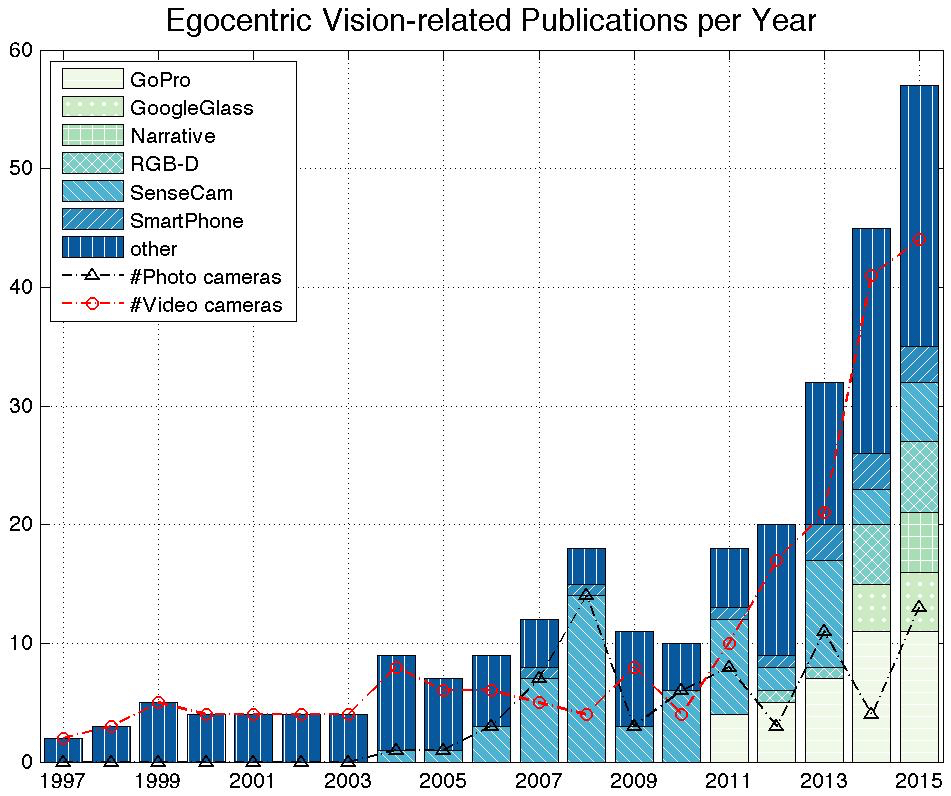}
	\caption{Histogram of the number of research papers published per year related to egocentric vision. The different colours indicate how many papers used each kind of camera. The dashed blue and black lines make a less specific distinction, showing the number of studies that used  photo (LTR) or video (HTR) cameras, respectively. 
}
\label{fig:year_graph}
\end{center}
\vspace{-1em}
\end{figure}
 




In response to the challenges and opportunities introduced by analysis of visual lifelogs, and more generally, by wearable cameras, computer vision scientists have rapidly become more interested in the subject over recent years.
By searching for the keywords  \emph{egocentric vision}, \emph{first person vision}, \emph{ego vision} and \emph{visual lifelogging}, using \emph{Google Scholar}, \emph{DBLP} and \emph{visionbib.com}, we found 274 papers in total devoted to visual lifelogging. For each of them, we annotated the type of camera used in the study and generated the plot in Fig. \ref{fig:year_graph}, which represents all the papers related to egocentric vision 
up to November 2015. As can be seen, interest grew very fast in the last years and the number of papers published increased by over 50\% in 2014 alone. Dotted lines show the comparatively small amount of work devoted to the analysis of image streams captured by photo cameras. This trend seemed to temporally change from 2007 to 2010, when the popularity of SenseCam resulted in a growth in the use of photo streams. 

An additional indication of the interest in this emerging field is the fact that in the last years, four surveys of wearable cameras and egocentric vision have been published. One, written by Doherty et al. \cite{Doherty2013Wearable}, focuses on explaining the \textit{ethical and data management issues} that must be taken into account when developing some health-related application using wearable cameras. The second one, by Betancourt et al. \cite{betancourt2014overview}, provides a general perspective on egocentric vision and devotes most of its analysis to the \textit{egocentric camera hardware}, \textit{egocentric datasets}, \textit{augmented reality}, \textit{algorithm types} and \textit{feature types} used in the literature from 1997 to 2014. This analysis is focused on providing a historical perspective of egocentric devices and their algorithms in addition to several ways of categorizing the existent papers in this field. The third one, which is a book by Gurrin et al. \cite{gurrin2014lifelogging}, focuses on data management and distinguishes between \textit{data storage}, \textit{organisation} and \textit{visualisation}; while also provides an overview of potential \textit{applications}. \textcolor{blue}{The fourth study, by Harvey et al. \cite{harvey2015remembering}, the authors present their work from the perspective of providing an aid to human memory.} They analyse the human memory mechanisms from a psychological perspective and propose a pipeline for enhancing it based on \textit{segmentation}, \textit{context enhancement (recognising objects and people)} and \textit{image retrieval}.

\textcolor{blue}{
This paper focuses on addressing the question: \textit{How far are we from being able to automatically tell our stories using egocentric photo streams?} The process of fully understanding the story behind the pictures is fundamental towards enabling a wide range of applications \cite{dimiccoli2015visual} and user cases \cite{hopfgartner2013user}, especially related to health.  As we explained, since these applications require observations over long periods of time, data should be acquired by photographic cameras (e.g. SenseCam, Narrative, etc.) instead of video cameras (e.g. GoPro, GoogleGlass, Looxcie, etc.). To this end, a thorough review of the published advances in egocentric data analysis is presented and research insights are provided. In contrast to previous surveys, we review and give details of studies that focus on both photographic and video cameras, considering which aspects should be reformulated and modified for their applicability in the LTR domain, and thus for egocentric storytelling.}

\textcolor{blue}{
To summarize, our contributions are as follows:
\begin{itemize}
	\item Review of methods for acquiring, organizing, summarizing and browsing large collections of unstructured data.  
	\item Organization of the available literature around the central questions necessary to address the storytelling problem: \textit{Was the user interacting  with somebody? How?}, \textit{Where} is he/she?, \textit{When} did the event occur? and \textit{What} is the person wearing the camera doing?.
	\item Highlights of the weaknesses and strengths of the reviewed techniques with respect to their applicability to the LTR domain (at the end of each subsection).
	\item Extensive analysis of the available datasets and source code related to the storytelling problems.
	\item Open problems and challenges in the field of egocentric vision with the final goal of storytelling.
\end{itemize}
}

The rest of the paper is organised as follows. In Section \ref{sec:partI}, we review the most important papers devoted to the task of acquiring, organising, summarizing and browsing large and unstructured collections of egocentric data. 
The solutions to these problems provide a basis to further analyse the data content, as in Section \ref{sec:partII}, where we review papers that claim to construct  semantic building  blocks for storytelling. 
Concluding remarks about applicability to the LTR domain are given at the end of each subsection. In Section \ref{sec:datasetsSoftware}, we summarize the available egocentric  datasets with the corresponding annotations, as well as the egocentric vision software.  Finally, in Section \ref{sec:conclusions}, we draw our conclusions and give some possible future directions for the research necessary to fill the gap between raw egocentric data analysis and visual storytelling.


\vspace{-0.5em}
\section{Visual lifelogging acquisition, segmentation and summarization}
\label{sec:partI}

This section reviews the literature concerning acquiring, structuring and summarizing visual lifelogging data,  which is summarized in Table \ref{tab:visual_management}.


\begin{table}[!ht]
  \centering
  \caption{Summary of all the visual lifelogging papers reviewed in this survey related to acquiring, organizing, summarizing and browsing large collections of unstructured data.}
  \begin{tabularx}{\columnwidth}{X X X X X X X X} \hline
    \multicolumn{8}{c}{TOWARDS STORYTELLING FROM VISUAL LIFELOGGING} \\						\hline
    
    \multicolumn{8}{c}{\ref{sec:salient_images} \textbf{Informative Images Detection}} \\	\hline
    \cite{xiong2014detecting} & \cite{lidon2015semantic} & & & & & & \\ 					\hline
    
    \multicolumn{8}{c}{\ref{sec:segmentation} \textbf{Temporal Segmentation}} \\				\hline
     \cite{li2013automatically} & \cite{Doherty:2008} & \cite{doherty2008combiningimage} & \cite{lin2006structuring} & \cite{spriggs2009temporal} & \cite{lu2013story} & \cite{bolanos2014video} & \cite{poleg2014temporal} \\ \cite{talavera2015r-clustering} & \cite{castro2015predicting} & & & & &	\\ \hline
     
    \multicolumn{8}{c}{\ref{sec:summarization} \textbf{Egocentric Summarization}} \\		\hline
     \cite{smeaton2010video} & \cite{jinda2012keyframe} & \cite{ghosh2012discovering} & \cite{lu2013story} & \cite{bolanos2015visual} & \cite{lidon2015semantic} & & \\														   \hline
     
    \multicolumn{8}{c}{\ref{sec:retrieval} \textbf{Content-Based Search and Retrieval}} \\	\hline
     \cite{wang2006vferret} & \cite{chandrasekhar2014incremental} & \cite{min2014efficient} & \cite{aghazadeh2011novelty} & \cite{wang2012semantics} & & & \\ \hline
  \end{tabularx}
\label{tab:visual_management}
\end{table}

\vspace{-1em}

\subsection{Data Acquisition}
The positioning of a wearable camera is of  crucial importance for  lifelogging data acquisition from the point of view of its later application. Mayol-Cuevas et al. \cite{mayol2009choice} evaluated, partially through simulations on a 3D facet model of the human body, four attributes of optical devices with respect to their position on the wearer's body:  social acceptability, absolute field of view (FOV), resilience to body motion, and view of the handling space region.
That study concluded that wearable cameras placed on the chest are the most socially acceptable and therefore offer the advantage of not interfering with social interactions. In addition, they are relatively resilient to the disturbances introduced by the wearer's own motion and are closely linked to the user's workspace, since they allow visualisation of the manipulative space in front of the wearer's chest. However, the FOV is quite narrow and does not allow the focus of the wearer's attention to be modelled.
In contrast, cameras worn on the head have a wider FOV and do allow this attention to be modelled, 
but they are the most sensitive to the wearer's motion and suffer from low social acceptability. A compromise between the size of the FOV, accessibility to the handling regions, sensitivity to ego-motion and social acceptability is offered by wearable cameras placed on the shoulder. The authors also considered the possibility of wearing multiple devices on different parts of the body so that their FOVs would be complementary, with the joint FOV computed as the union of the individual FOVs. 

\textit{Remarks:} 
Since for long-term image acquisition social acceptability is crucial, placement on the chest is usually considered the best choice. In addition, it has the advantage of offering access to the handling space and the  manipulation of objects can be focused.

\subsection{Informative Image Detection}\label{sec:salient_images}

Once images have been acquired, before proceeding with any structuring, analysis and summarization, proper cleaning of the images is necessary. This need stems from the fact that egocentric images are non-intentional images, that is, nobody decides when and of what to take a picture. As a result, a significant number of images can be blurred, can be dark, or can capture non-informative data (the sky, the ground, walls, etc.). In Xiong and Grauman \cite{xiong2014detecting}, informative images are defined as "intentional" images, obtained once those  with undesired artefacts, such as light saturation,  blurred images, or useless information (the sky, walls, etc.) have been removed. Lidon et al. \cite{lidon2015semantic} define as informative any image that includes objects and/or people, and which is of reasonable quality, assuming that it does not include any undesired artefacts (e.g. blurring, darkness or occlusions). With this definition, they trained a binary CNN to make this distinction.  

\subsection{Temporal Segmentation}\label{sec:segmentation}

\begin{figure}[!t]
	\vspace{-1em}
    \centering
    \includegraphics[width=0.9\columnwidth]{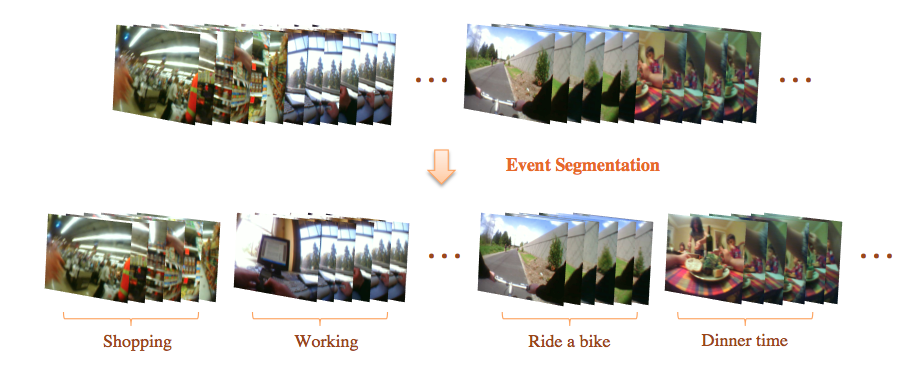}
  \caption{Example of the desired event segmentation applied to lifelogging data. The goal is to group with respect to their main event, considering the activities, objects or people involved.}
  \label{fig:event_segmentation}
\end{figure}

Lifelogging data typically consist of long unstructured videos or photo streams. Organising and structuring them into homogeneous temporal segments, corresponding to different events and/or environments (see Fig. \ref{fig:event_segmentation}), are very important to facilitate browsing and analysis of the images. State-of-the-art methods for egocentric data segmentation  can be classified into two broad classes depending on whether the homogeneous segments represent  what the {\em wearer sees} or  
{\em does}. 

The former class uses features that can capture the characteristics of the environment around the wearer as image representation. Early work aiming at segmenting the sequences into visually homogeneous segments was based on low-level features. Li et al. \cite{li2013automatically} have proven that it is possible to distinguish different events simply by treating SenseCam images as time-series data and calculating the eigenvalue peaks in consecutive windows of images. Doherty et al. \cite{Doherty:2008,doherty2008combiningimage} used different descriptors for image representation and the metadata available from the camera sensors. Lin and Hauptmann \cite{lin2006structuring} proposed a simple approach based on using colour features in a time-constrained K-means clustering algorithm, capable of maintaining temporal coherence on the  splitting of events. Spriggs et al. in \cite{spriggs2009temporal} proposed a method for simultaneous temporal segmentation and recognition of activity related to cooking. They captured videos at the same time from a single wearable video camera and multiple other static cameras, sensors, microphones, etc., and used both sensor data and visual GIST descriptors to describe the frames. For the unsupervised scene segmentation, they applied a Gaussian mixture model. More recently, Talavera et al. \cite{talavera2015r-clustering} proposed the use of CNNs computed on the whole image using AlexNet as a fixed feature extractor for image representation. That work, designed for egocentric photo streams, uses a graph-cut algorithm to temporally segment the photo streams and includes an agglomerative clustering approach with concept drifting methodology, called ADWIN.

\begin{figure}[h]
	\vspace{-0.5em}
    \centering
    \includegraphics[width=0.8\columnwidth]{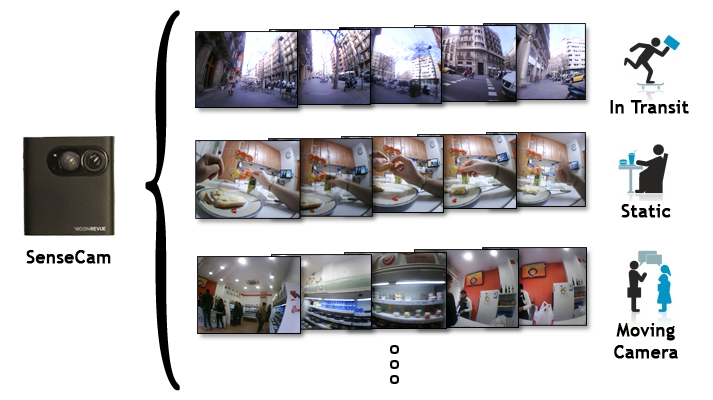}
  \caption{Motion-based segmentation framework proposed in \cite{bolanos2014video}. By including motion features to describe egocentric pictures they can separate the events considering the dynamism of the activities performed.}
  \label{fig:motion_segmentation}
   \vspace{-2em}
\end{figure}

Methods focusing on what the camera wearer does mostly use motion information as image representation.
Usually, optical flow is used to distinguish between {\em static}, {\em moving the head/camera} and {\em in-transit} frames \cite{bolanos2014video,lu2013story} (see Fig. \ref{fig:motion_segmentation}). 
To focus on long-term ego-activities, Poleg et al. \cite{poleg2014temporal} proposed the use of so-called integral motion, which is closely related to the wearer's activity. By integrating the instantaneous displacements at fixed image patches, the variations due to head rotation are eliminated, since their mean is practically zero, leaving only the consistent displacement caused by forward motion. 
A different approach, based on CNNs, is adopted by Castro et al. \cite{castro2015predicting}. They gathered a large egocentric dataset from a single user and fine-tuned a CNN pre-trained on ImageNet for activity classification. They proved that the  network trained on the data of a single user can be re-trained to generalise to new users. The main problem with this approach is that a new set (several thousands of images) must be labelled from scratch whenever it is necessary to predict the events affecting a new user with the model. 


\textbullet\hspace{0.5em}\textit{Remarks:}  The applicability of motion as a feature, though relevant when dealing with videos, has proven to be rather limited for photo streams. In the latter case, the use of richer representations, such as global CNN-based features, seems crucial to compensate this limitation. The use of time-dependent methods for egocentric segmentation is also a must considering the nature of the data. A promising approach to improve the results of the segmentation of egocentric sequences is the addition of semantic-level features (scenes, objects, social interaction, actions, etc.). This additional information would be an important step to bring machine segmentation closer to the way humans segment unconstrained streams of images.

\subsection{Egocentric Summarization}\label{sec:summarization}

Summarization is the process of generating a proper, compact and meaningful representation \cite{truong2007video} of a given sequence through a subset of representative frames or segments. This step is crucial to help manage and browse large volumes of lifelogging video content efficiently. Basically, there are two kinds of summaries that can be produced: a static video story board, which is composed of a set of salient images extracted or synthesised from the original sequence, and dynamic video skimming: a shorter version of the original video made up of several shots, comprised of a series of frames. 
To fully exploit the potential of visual lifelogs in a variety of applications, an egocentric summarization method should be designed to aid in the visualisation, indexing and browsing of autobiographical events, with the least possible semantic loss. 

Story board summarization has been traditionally formulated as grouping images into coherent collections by relying on low-level spatio-temporal features and then selecting the most representative image (or set of images) from each collection \cite{smeaton2010video}. Based on this classical approach, Jinda-Apiraksa et al. \cite{jinda2012keyframe}  and Chowdhury et al.  \cite{Chowdhury2015My}, developed similar techniques for keyframe selection in egocentric sequences based on quality measures \cite{jinda2012keyframe,Chowdhury2015My} and both quality and diversity measures \cite{Chowdhury2015My}. More complex features for grouping were used by Bola\~nos et al. \cite{bolanos2015visual}. Their methodology, adapted for photo cameras, uses the AlexNet CNN as a feature extractor to characterise each frame. Then, using those features, they apply event segmentation using a hierarchical clustering algorithm and a posterior single keyframe selection by applying the Random Walk algorithm to each of the segments.



While these methods rely solely on low-level features, some recent work has introduced a semantic level in the keyframe selection process. Ghosh et al. \cite{ghosh2012discovering} suggested that video summarization should be driven by the presence of important people and objects. Following this idea, they proposed a method that reveals salient people and objects based on their interaction time with the camera wearer and then selected keyframes according to keyobject event occurrences. Lu and Grauman \cite{lu2013story}, following on from their previous work, suggested that video summarization should preserve the narrative character of a visual lifelog and proposed a shot selection consisting of three terms: 1) a term that models \textit{story coherence} by favouring shots capable of following the inherent story; 2) a term that models \textit{importance}, to choose only shots that show some important aspect of the day; and 3) a term that models \textit{diversity} and avoids repeating similar events.

Summarization that considers semantic topics was recently proposed by Varini et al. \cite{Varini2015Per} and Schinasi et al. \cite{Schinasy2015Vis}. In  \cite{Varini2015Per},   it is assumed that interesting scenes in a cultural experience, such as visiting a museum, are those  associated with certain patterns of behaviour of the camera wearer that are learned and used for classification. Taking into account the topic of interest of the user, different summaries can be generated from the same video. 
In \cite{Schinasy2015Vis}, topics are revealed from a set of social media messages as highly connected messages in a graph, whose nodes encode messages and whose edges encode their similarities. Finally, the images that best represent the topic are selected based on their relevance and diversity. Lidon et al. \cite{lidon2015semantic}, also working on photo sequences, proposed an event keyframe ranking method based on a trade-off between image relevance and diversity after removing non-informative images (containing undesired artefacts, e.g. blurring, darkness or occlusion, or showing the sky, walls or object parts) by using a new binary CNN-based filter. Their relevance criteria took into consideration several semantic measurements, including whether faces and/or objects were present, as well as whether the images had a high saliency value. 

\textbullet\hspace{0.5em}\textit{Remarks:} A semantic-oriented approach to egocentric summarization seems to be the most suitable for lifelogging data. Indeed, users would ideally search for complex autobiographical events that encompass simpler human actions and may not be directly correlated with their visual appearance.
When dealing with photographic cameras, and due to the nature of their data, the only possible way to tackle the summarization problem is through the keyframe selection approach. Taking this into account, methods like \cite{lu2013story} should be reformulated, either considering the video sub-shots as single frames, or developing a fine-grained segmentation procedure. This procedure should separate the data into a large number of events to have enough segments to apply the sub-shot selection correctly.

\subsection{Content-Based Search and Retrieval}\label{sec:retrieval}

Retrieving images from a large personal database allows us to browse, search and find images of previously seen objects or places and thereby has the potential to solve a broad range of problems in egocentric vision, such as:
\begin{itemize}
	\item searching for elements (Have I seen this before?);
	\item navigating (How often do I visit this place?);
	\item understanding the environment (Where am I right now?);
    \item efficiently organising huge amounts of data.
\end{itemize}



Following these premises, in \cite{wang2006vferret} Wang et al. built a system for content-based searching and browsing that starts by splitting the stored data into segments and extracting three kinds of information: 1) time and other relevant attributes, 2) low visual features, and 3) audio features.  Then, in the retrieval step, they applied time-based filtering by comparing the time attributes of the images in the database with the query introduced by the user. A clustering step then extracts a representative clip from each cluster; and finally, the user can provide one or more query images for the system to refine the search based on visual features and improve the query result. Still, several open issues remain: in many situations it is difficult to recall the time and where the photo we are looking at was taken; visual features are too simple to capture real object shape and texture differences; and furthermore, audio features are not provided by all wearable devices. 
Aghazadeh et al. \cite{aghazadeh2011novelty} proposed to retrieve  novel scenes and actions with respect to a previously acquired egocentric  dataset by using a set of "alignment" sequences, and matching them with a new "query" sequence by using dynamic time warping.

Assuming that searching, browsing
or summarization in visual lifelogging would largely benefit from semantic concept representation,  Wang and Smeaton \cite{wang2012semantics} investigated the selection of the most appropriate combination of concepts for event representation. Their strategy basically consists of reasoning on semantic networks using a density-based approach.
Min et al. \cite{chandrasekhar2014incremental,min2014efficient} represented millions of egocentric images on a sparse graph. They represented each image as a node in the graph, and added an edge between two nodes, when they belonged to the same bag in a BoW representation. Relying on this representation, they showed that local density clustering is more suitable than global clustering methods, considering the high redundancy that lifelogging data inherently possess.


\textbullet\hspace{0.5em}\textit{Remarks:} Many issues remain regarding content-based retrieval techniques, for instance: How can we make use of the basic building blocks extracted from lifelogging (actions, people and environments)? The usage of a multi-level and multi-modal descriptions based on the recognition of actions, people, objects and environments could provide a detailed image description close to text-level, which could allow high retrieval accuracy.

In methods such as \cite{chandrasekhar2014incremental,min2014efficient}, new challenges would arise when dealing with photo data, considering the higher variability of consecutive images compared to video sequences.

\section{Visual Lifelogging Analysis}
\label{sec:partII}

We present an overview of the most important papers on visual lifelogging analysis and the problems they tackled, organised around four basic questions: \textit{Is the user interacting? How?} \textit{Where} is the user? \textit{When are the events occurring?}  and \textit{What} is the user doing?. Table \ref{tab:visual_analysis} lists the papers and related information.

\begin{table}[!ht]
  \centering
  \caption{Summary of all the visual lifelogging analysis-related papers reviewed.}
  \begin{tabularx}{\columnwidth}{X X X X X X X X X} 										\hline
    \multicolumn{9}{c}{VISUAL LIFELOGGING ANALYSIS} \\									\hline
    
    \multicolumn{9}{c}{\ref{sec:who} \textbf{Interacting? How?: Social Interactions}} \\		\hline
    
     \cite{doherty2008combining} & \cite{alletto2014head} & \cite{alletto2014ego} & \cite{fathi2012social} & \cite{bazzani2013social} & \cite{aghaei2015multi} & \cite{aghaei2014bag} & \cite{aghaei2015towards} & \cite{poiesi2014predicting} \\ \cite{soo2015social} & & & & & \\ \hline

    \multicolumn{9}{c}{\ref{sec:where} \textbf{Where?: Scene Understanding}} \\			\hline
    
    \multicolumn{4}{c}{Concept Recognition} & \cite{byrne2010everyday}   \\																								\hline
    
     \multicolumn{4}{c}{\ref{subsec:obj_det_rec_dis} Object Recognition} & \cite{ren2009egocentric} & \cite{ren2010figure} & \cite{fathi2011learning} & \cite{bolanos2013active} & \cite{bettadapura2015leveraging} \\											\hline
     
    \multicolumn{4}{c}{\ref{subsec:obj_det_rec_dis} Object Discovery} & \cite{kang2011discovering} & \cite{bolanos2015object} & \cite{bolanos2015ego-object} &   \\ 						\hline
    
    \multicolumn{4}{c}{\ref{subsec:spatial_localization} Spatial Localisation} & \cite{kikhia2014structuring} & \cite{bettadapuraegocentric} & \cite{wannous2012place} &  \\ 							\hline
    
    \multicolumn{9}{c}{\ref{sec:when} \textbf{When?: Time-Based Localisation}} \\				\hline
    \cite{lin2006structuring} & \cite{talavera2015r-clustering} & \cite{castro2015predicting} & & & & \\ \hline
    
    \multicolumn{9}{c}{\ref{sec:what} \textbf{What?: Action Recognition}} \\				\hline
    
    \multicolumn{4}{c}{\ref{subsec:body_movements} Body movements} & \cite{poleg2014temporal} & \cite{kitani2011fast} & & \\		\hline
    
     \multicolumn{4}{c}{\ref{subsec:object_hand_interaction} Object-hand interaction} & \cite{fathi2011understanding} & \cite{sundaram2010egocentric} & \cite{behera2013egocentric} & \cite{pirsiavash2012detecting} \\ & & & & \cite{li2013pixel} & \cite{lee2014hand} & \cite{rogez2014egocentric} & \cite{rogez2014_3d} \\																\hline
     
    \multicolumn{4}{c}{\ref{subsec:attention_based} Attention} & \cite{fathi2012learning} & \cite{matsuo2014attention} & \cite{ryoo2013first} &  \\ 							\hline
    
    \multicolumn{4}{c}{\ref{subsec:other_approaches} Other Approaches} & \cite{yan2014recognizing} & \cite{song2014activity} & \cite{kanade2012first} & 	\\ 
    \hline
    
  \end{tabularx}
  \label{tab:visual_analysis}
\end{table}

\subsection{Interacting? How?: Social Interactions}\label{sec:who}

Following the definition by Rummel \cite{rummel1976social}, \textit{social interactions} are all acts, actions or practices of two or more people mutually oriented towards each other. 
Given the powerful social nature of humans, the analysis of social interactions in lifelogging data is of fundamental importance to understanding human behaviour. Furthermore, the presence of people and social interactions are consistently associated with event memorability \cite{isola2011makes} and therefore, their detection is also potentially useful for keyframe extraction or to estimate the importance of events in a lifelog \cite{doherty2008combining}. From the perspective of computer vision, social interactions can be characterised by patterns of attention between individuals. Analysing attention patterns requires the detection, tracking and locating of people in 3D environments. Indeed, when interacting with others, we naturally tend to place ourselves in certain positions so as to stand close to those we interact with and avoid occlusions. F-formations \cite{kendon1977studies} have been demonstrated to be a suitable formalism for modelling social interaction behaviour. Following the original definition by Kendon \cite{kendon1990conducting}:
\begin{displayquote}
\textit{An F-formation arises whenever two or more people sustain a spatial and orientational relationship in which the space between them is one to which they have equal, direct, and exclusive access.}
\end{displayquote}

\begin{figure}
\begin{center}
\includegraphics[width=0.8\columnwidth, height=0.7\columnwidth]{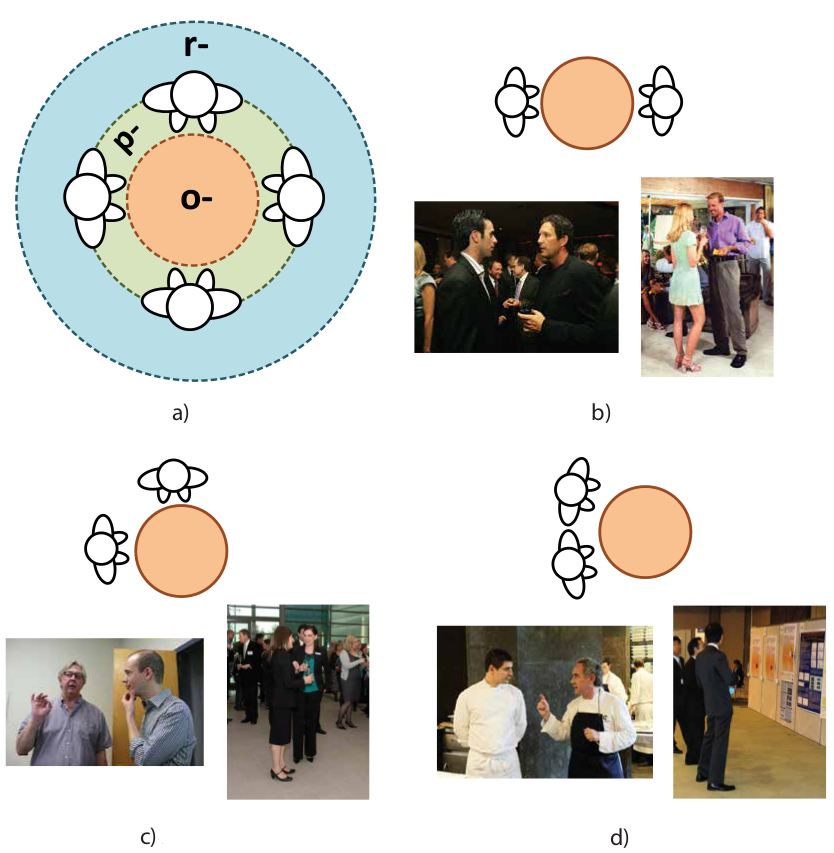}
\caption{Different arrangements of F-formations that are useful for social interaction analysis: (a) circular arrangement. (b) vis-a-vis arrangement. (c) L-arrangement. (d) side by side arrangement. Image adapted from \protect\cite{setti2014f}.}
\label{Fformations}
\end{center}
 \vspace{-0.5em}
\end{figure}

\begin{figure}[!ht]
\centering
\includegraphics[width=0.8\columnwidth]{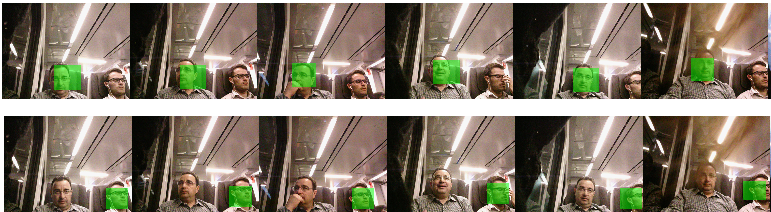}
\caption{Example of multi-face tracking obtained by applying the method in \protect\cite{aghaei2015multi} to track multiple faces in LTR sequences captured by a wearable camera. Each row represents the track of a different person.}\label{maya}
\vspace{-1em}
\end{figure}

Examples of F-formations are given in Fig. \ref{Fformations}. The F-formations theory has been successfully applied in social interaction analysis \cite{hung2011detecting} using classical videos or still images, and more recently to egocentric videos \cite{alletto2014ego}. Head estimation and 3D location are crucial for the detection of F-formations. Indeed, a rough estimate of someone's head pose allows us to understand with a certain precision what the person is looking at; 
while it is important to estimate the distance people have from the camera wearer and other people if there is interaction.

In sequences captured through a wearable camera, pose estimation is a challenging task due to the continuous  changes of aspect ratio, scale and orientation. A common way to address this problem \cite{alletto2014head,alletto2014ego,bazzani2013social,poiesi2014predicting} is to assume that where a group interacts in a discussion, the head of each person will be oriented for a while towards the person who is speaking, and to use a model to capture this behaviour over time. Generally, in video sequences, this is achieved through a hidden Markov model or Markov random fields, where the latent variable corresponds to the head pose and the observed variables to the results of a multiple person tracker, applied to the input images. The only works devoted to the analysis of photo sequences are \cite{aghaei2015multi, aghaei2014bag, aghaei2015towards}. In this context, tracking people is very challenging due to the abrupt and very frequent changes of view. The proposed approach basically consists of computing backward and forward correspondences for each face detected in the sequence and of grouping similar tracklets into bags, which should correspond to different people (see Fig. \ref{maya}).
A combination of first-person and third-person views is considered by Soo and Shi \cite{soo2015social} to predict social saliency, considered as the likelihood of joint attention, in real-world scenes with multiple social groups. This is basically achieved by modelling social formation features that encode the geometric relation between the joint attention and spatial distribution of the members of a social group.   
\textbullet\hspace{0.5em}\textit{Remarks:} In general, there is common agreement about the need to track people, head orientation  and 3D locations to detect F-formations that represent social groups in egocentric sequences; however, two fundamental problems arise. First, since in different social scenarios, distances and poses can assume different degrees of significance, clearly a need emerges for an algorithm to be able to adapt to different situations and learn how to treat distance and orientation features depending on the context. As a consequence, the choice of which data to use for training is crucial. Second, distances and poses strongly depend on where the camera is worn (eyeglasses, on the head, on the neck, etc.). Except \cite{aghaei2015multi, aghaei2014bag}, all the methods mentioned above  rely strongly on temporal coherence, since they were conceived for video sequences. Further advances in the analysis of social interactions through photographic cameras would require us to focus on features that are less sensitive to changes over time, such as people's body movements, which are consistently associated with emotional experiences \cite{mehrabian1969significance} and could, therefore, be considered cues of social interactions.


\subsection{Where?  Scene Understanding}\label{sec:where}

To answer the question "Where is the user?", we require a semantic understanding of the elements that surround the camera wearer, such as  objects, people and environments, since they represent the cues available to recognise his/her surroundings. In this section, we provide an overview of computer vision tasks related to scene understanding, such as \textit{object recognition}, \textit{spatial localisation}, \textit{scene parsing} and \textit{scene recognition}. All of them share the goal of determining what the most promising techniques are for understanding scenes in lifelogging data.


\subsubsection{Object Recognition and Object Discovery}\label{subsec:obj_det_rec_dis}


Scenes can also be characterised by a vocabulary of concepts that can be found in them. With this aim,  we consider the following problems: \emph{object recognition}, which intends to identify the category that a given object belongs to; and \emph{object discovery}, which detects, recognises and reveals new objects in images  that possibly have never been seen before by the algorithm in the previous images. Due to the free motion of the camera and to the passive acquisition of lifelogging data, objects are frequently occluded and their appearance may vary broadly. Thus, the \emph{object recognition} problem in egocentric data is becoming a challenging and active research field. The first work on object recognition in the domain of lifelogging is by Byrne et al. \cite{byrne2010everyday},  who successfully validated  supervised concept recognition, referring  to relevant objects or scenes as concepts. Furthermore, using the output of the detector, they showed that the images that compose a lifelog collection tend to be temporally consistent in their visual properties, as well as in the concepts they contain. Because of this concept consistency, they suggested that an efficient automatic extraction and inference of higher-level semantic concepts based on co-occurrences and known relationships would be feasible. Bola\~nos et al. \cite{bolanos2013active} developed an active labelling method to generate a sufficiently large number of training examples to train an efficient supervised classifier. The method, based on a combination of hierarchical clustering trees, uses an unsupervised learning algorithm to organise the data, selecting the most informative part, asking the user for their labels, and using the feedback provided to improve the classification in a semi-supervised way. Ren et al. in \cite{ren2009egocentric,ren2010figure} and  Fathi et al. in \cite{fathi2011learning} used head-mounted  cameras and proposed methods that recognise objects held in the user's hand. 
They segmented the background from the foreground (hands and objects) using optical flow features and relying on the fact that foreground objects will usually move in a more dynamic way while the background is more static.

\begin{figure}[!ht]
    \centering
    \includegraphics[width=0.9\columnwidth]{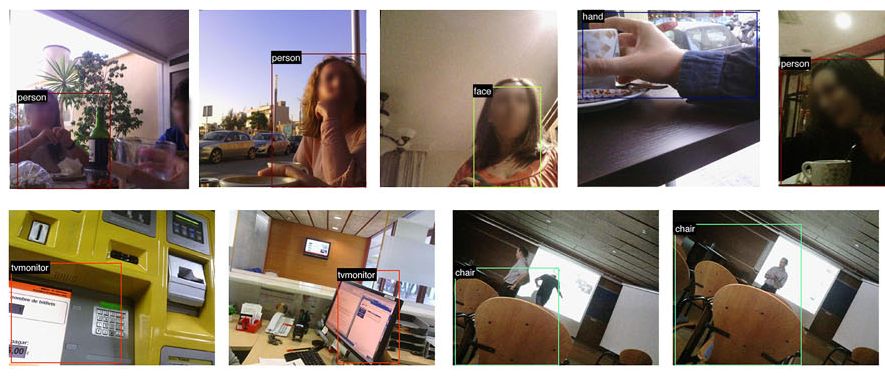}
  \caption{Examples of objects revealed by the ego-object discovery methodology \cite{bolanos2015ego-object} for two different subjects (one per row). Better viewed in digital format.}
  \label{fig:ego-objects}
\end{figure}

Focusing on the task of \emph{object discovery} in lifelogging data (see example in Fig. \ref{fig:ego-objects}), Kang et al. \cite{kang2011discovering} proposed a method, starting from an initial segmentation, that clusters only samples with higher correlation that should belong to the same object type. 
To this end, starting from the initial segmentation, they provide a merging strategy for segments that closely co-occur in most images. In this way, they complete objects that might be composed of  different, but clearly defined parts (e.g. a laptop composed by a screen and keyboard). 
With the same goal, Bola\~nos et al. in \cite{bolanos2015object,bolanos2015ego-object} proposed the use of a state-of-the-art objectness detector and a pre-trained CNN specialised in object recognition to extract a set of rich features for each object candidate followed by clustering them. The clustering integrates a "Bag of Refill" strategy of previously discovered object instances as a knowledge reuse methodology.

\subsubsection{Spatial Localisation}\label{subsec:spatial_localization}

\begin{figure}[!ht]
\begin{center}
\includegraphics[width=0.7\columnwidth]{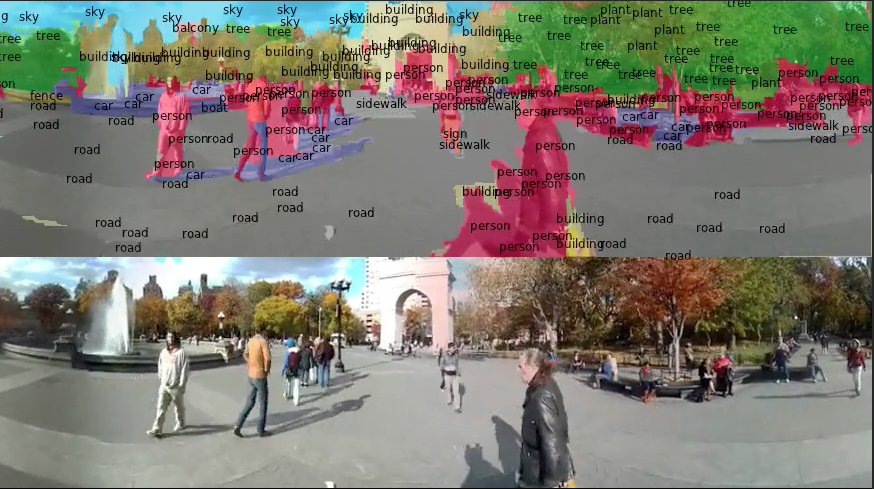}
\caption{Example of the result obtained (top) by applying a scene parsing algorithm to a conventional non-egocentric image (bottom). We can see the different segments found (separated by different colours) and the classes assigned to each of them. Picture adapted from \cite{farabet2013learning}.}
\label{fig:scene_parsing}
\end{center}
 \vspace{-0.5em}
\end{figure}

Bettadapura et al. in \cite{bettadapuraegocentric}, proposed a method called FOV localisation that combines localisation techniques  with egocentric images to localise the user(s) in the environment. To do so, they used a reference dataset, which can be images from Google Street View or pre-recorded videos from fixed cameras, and matched them to the data acquired by the user's photographic or video camera to obtain his/her localisation. 
They tested the system on multiple datasets captured indoors and outdoors. Additionally, they proposed a combined FOV localisation system for simultaneous localisation of multiple users of wearable devices. Wannous et al. in \cite{wannous2012place} also proposed a methodology for localisation and action-related event recognition. They used a shoulder-mounted video camera to acquire images of daily indoor living (e.g. kitchen, office, library, etc.) and built a 3D model of the different scenes. In their work, they proved that their models were more powerful than simpler 2D ones and were able to recover information from previously seen scenes with query images.



\textbullet\hspace{0.5em}\textit{Remarks:} Another interesting approach that egocentric vision could benefit from is scene parsing. This is based on image segmentation; that is, separating out all the regions in an image that belong to different objects or regions. Furthermore, these kinds of techniques classically consist of providing pixel-level segmentation of the whole image and at the same time assigning an object class to each of the pixels (see the example of scene parsing in Fig. \ref{fig:scene_parsing}). To do this, most of the methods use pixel-level classifiers to achieve an initial segmentation and then a graphical model is applied to smooth and correct the boundaries of the segments \cite{farabet2013learning,yang2014context}. A limited amount of work in this field can be found in the literature but none of it was specifically designed or tested on egocentric and lifelogging datasets.
Considering the differences we could find in an egocentric dataset (and more precisely in lifelogs) with respect to those typically used in scene parsing, we can enumerate some clear points to take into account when working on scene parsing:

\begin{itemize}
  \item  Scene parsing datasets are usually composed of natural and urban scenes (in general, outdoors) and their corresponding class distributions have a high percentage of training samples related to those environments, that is, the egocentric lifelogging datasets for scene parsing would be very different considering the indoor and routine settings where people usually spend most of their time. 
  \item Also taking into account the fact that egocentric vision datasets are composed of routine and redundant scenes, scene parsing methods focusing on lifelogging images should provide some higher context and knowledge-reuse mechanisms to take advantage of the previously parsed images in the egocentric sequence.
\end{itemize}

Related to scene parsing, it would also be useful to be able to recognise the scene the user is in. Although no work has been presented with this purpose using egocentric data, a good example with conventional images is the dataset Places205 \cite{zhou2014learning}. This information could help when deciding, for instance, how we should segment the day into events or use this information to exploit the environment-object relationships.

Although good methodologies have been proposed for object recognition and object discovery using egocentric and lifelogging images, there is still a lot of work to do to semantically describe the camera wearer's environment at a high level. The development of object detection methods specifically designed for egocentric images could not only improve existent recognition and discovery methods, but also set a more robust basis for the future appearance of scene parsing of lifelogging images. To achieve these goals, new computer vision techniques able to cope with blurring, light saturation  and  the occlusion of objects have to be developed. Hence, new techniques for gathering huge labelled datasets not only for object detection, but most importantly for scene parsing, must be developed. Furthermore, the addition of GPS or visual localisation techniques to scene parsing could clearly improve understanding of the environment. The most promising technique applicable to scene parsing is using Fully Convolutional Networks \cite{long2014fully}, which are able to infer the classes of each pixel treating the image as a whole instead of the current pixel-level centred classifications.

Finally, note that all the work on object recognition relies on the user-like focus and point of view that head-mounted cameras offer. This approach would not be feasible for real applications, where neck hanging cameras are usually used because they are considered less obtrusive and more user-friendly \cite{hayden2012wearable}, despite not always being able to show what the user is doing. Moreover, these algorithms, which rely heavily on temporally close video frames and motion information, would not be applicable to LTR photographic cameras either.

\vspace{-0.5em}

\subsection{When? Time-Based Localisation}\label{sec:when}

Time information is particularly important to determine the causal relations in human behaviour. For instance, it could be useful in understanding which factors determine crises in people affected by bipolar disorder. The most common annotation tool used for keeping a record of the time in lifelogging data is the time stamp provided by cameras. 
By using this information, one can easily establish the temporal placement of the data in the long term, the order of the images, and their temporal distance for photographic cameras in the short term or daily. Some works have studied incorporating temporal information as a complementary feature indicator for achieving an indirect prediction. As an example, in \cite{lin2006structuring, talavera2015r-clustering} the authors have treated the data acquired as a time-series  to properly segment the different events present in a day. In \cite{castro2015predicting}, both the day of the week and the time of the day have been used for training a classifier with the ability to categorise different events.
Naaman et al. \cite{naaman2004context} studied the role of the time stamp as a memory cue in a psychological experiment on conventional images and concluded that people are unable to retrieve their memories when only given the time and date; consequently, additional information is needed for retrieval methods to be effective.


\subsection{What? Action Recognition}\label{sec:what}

\begin{figure}[ht]
	\vspace{-0.5em}
    \centering
 \includegraphics[width=0.8\columnwidth]{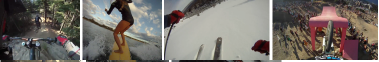}
  \caption{Examples of first-person point of view images performing various sports. Image adapted from \protect\cite{kitani2011fast}.
  }
  \label{fig:sports}
\end{figure}

Inferring what the camera wearer is doing from a visual lifelog basically requires the categorisation of everyday activities. The categories to focus on depend on the kind of application. For instance, in healthcare and well-being applications, occupational therapy research may guide the selection of the target activities and related concepts (see Fig. \ref{fig:sports} as an example of sports category recognition). For diet monitoring applications, eating actions will be the focus; whereas  in applications related to the diagnosis of dementia, the focus will be on daily life activities such as dressing, making coffee and cooking. In quantified-self applications, activities like  housework, watching TV, working/studying, eating/drinking, etc. are the most prevalent activities. 

Traditional action recognition methods can be broadly classified depending on the kind of features they use to represent actions; with  body movement analysis and the use of the objects involved in the action being the most common choices. Only very recently has the scene context  been used to improve action recognition. Still, the choice of the representation strongly depends on the kind of actions to be classified. 

\subsubsection{Body movement-based methods}\label{subsec:body_movements}

In an egocentric setting, general body movements such as running, walking, moving the head/camera or staying still are usually estimated relying on motion features (when this is possible with the temporal resolution of the camera). Usually, based on such features, the ego-action classification can also be used for event segmentation. Typically, video cameras like GoPro, which capture around 30 fps, are used to gather data. Poleg et al. \cite{poleg2014temporal} proposed integrating instantaneous displacements of fixed image patches over a long period of time  to remove the zero mean variations due to head rotation. By applying this process, they leave only the consistent displacement caused by forward motion. The cumulative displacement curves  show different patterns for  ego-motion activities, so that activities become easy to classify. Instead of focusing on the goal of building discriminative motion features, Kitani et al. \cite{kitani2011fast} used several modifications of classical motion-based feature vectors and built a complex Bayesian model for clustering.

\subsubsection{Object-hand interaction-based methods}\label{subsec:object_hand_interaction}

\begin{figure}[h]
	\vspace{-0.5em}
    \centering
 \includegraphics[width=0.7\columnwidth]{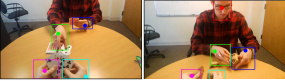}
  \caption{Examples of first-person point of view images for recognising activities involving hands. The algorithm is capable of detecting the left and right hand of the user, in pink and light blue respectively; and the left and right hand of the person he/she is interacting with, in dark blue and green, respectively. Image adapted from 
\protect\cite{lee2014hand}, devoted to hand disambiguation.}
  \label{fig:hand_disambiguation}
  \vspace{-0.5em}
\end{figure}

A first-person point of view offers an ideal perspective from which to analyse hand-object manipulation or hand-eye coordination (see Fig. \ref{fig:hand_disambiguation}). 
The main idea, introduced by Fathi et al. \cite{fathi2011understanding} and further improved by \cite{behera2013egocentric,pirsiavash2012detecting,sundaram2010egocentric} is that objects are correlated with actions (e.g. dish and nibbling) and actions with activities, and these correlations can be exploited to build robust object models.
However, the challenges come from additional occlusions (from manipulated objects, or self-occlusions of fingers by the palm) and the fact that hands interact with the environment and often leave the camera FOV. 

Others have focused on different problems related to hand-object manipulation such as capturing the variability of hand appearance over a diverse set of imaging conditions and hand poses \cite{li2013pixel}, disambiguating and tracking the observer’s hands and those of social partners \cite{lee2014hand}, improving robustness against camera motion \cite{ren2010figure,rogez2014_3d,rogez2014egocentric}, or capturing the appearance of visual composites of humans and objects in interaction \cite{pirsiavash2012detecting}. 

\subsubsection{Attention-based methods}\label{subsec:attention_based}

The use of manipulation-based approaches is restricted to scenes and objects where the user's hands present significant information. Attention-based approaches aim to identify objects to which the user pays particular attention, even in the absence of manipulation, since they could be key factors in self-behaviour recognition. In general, these methods are applicable to data acquired by head, eyeglass or ear-mounted cameras only.
Attention can be used to find salient objects as in Matsuo et al. \cite{matsuo2014attention}, or to capture the relationship between action and gaze, as in \cite{fathi2012learning}. 




\subsubsection{Other approaches}\label{subsec:other_approaches}

To detect activities that cannot be fully characterised by body movement, object-hand manipulation or object-gaze relationships, motion has been the most commonly used feature. 
Instead of trying to compute ego-motion, these approaches describe the frames that compose the actions, they use a set of motion and visual word features 
in a local (on a single frame) and global (on a set of consecutive frames) manner and create a specific structure for obtaining a temporally and spatially consistent representation of the action.
Song et al. \cite{song2014activity} obtained an accuracy rate of activity recognition of about $80\%$ using the dataset they published (LENa dataset), by adopting the dense trajectory approach.
In \cite{ryoo2013first}, the authors used a wearable video camera to capture and recognise a diverse set of actions (e.g. throwing, hand shaking, hugging or waving) which, in this case, is made by other people towards the camera user.
Recently, a newer approach for action recognition was proposed by the same authors in \cite{ryoo2014pooled}. On this occasion, they used CNN features to describe the frames of an HTR video. To obtain a rich and motion-like representation, they then proposed the use of a temporal pooling operator (PoT).
An interesting alternative to motion was proposed by  Yan et al. \cite{yan2014recognizing}, who exploited the fact that typically people tend to perform the same actions in the same environment (e.g. people at work typically have a coffee break) and their results  show the advantage of sharing information between tasks.
Kanade et al. \cite{kanade2012first} explored the problem of activity recognition from a deeper perspective. They proposed several methods for activity recognition, some based on object and scene understanding, which are specifically adapted to their eye-glass mounted wearable device.


\textbullet\hspace{0.5em}\textit{Remarks:} In essence, the most common cues on which activity recognition in egocentric videos relies on are body movement, object-hand interaction and patterns of attention. Body movement-based methods rely on motion estimation and therefore are not directly applicable to data acquired by photographic cameras. Object-hand interaction and patterns of attention are feasible for data acquired by wearable cameras attached to the head or somewhere near the person's eyes that could follow his/her gaze. However, when the camera is worn as a necklace or attached to the clothes, attention-based methods fail, making it impossible to see what the user is manipulating and making it very difficult to estimate the centre of attention. Similarly, object-hand methods can be very difficult to apply considering the free motion of the camera and the difficulty in regularly showing the hands of the user. To the  best of our knowledge, there is no published work on recognition of egocentric activities recorded by freely worn cameras. In this context, it would be a requirement for robust activity recognition to take into account information concerning whether the camera wearer is stationary or moving.

\section{Availability of datasets and software}\label{sec:datasetsSoftware}

\subsection{Egocentric Vision Datasets}\label{sec:datasets}

As egocentric vision is a relatively new research field, the creation of standardised and rich enough datasets and annotations to test and compare the new algorithms is crucial to boost the development of the field. 
In Table \ref{tab:datasets}, we provide a summary of currently available public egocentric datasets, specifying,  for each of them, the following information: the name and the reference paper where the datasets were presented or were used for the first time (where data can be found); a short description; the kind of annotated data they contain; and the camera used to acquire the data.

Only two of the publicly available egocentric datasets, EDUB \cite{bolanos2015ego-object} and AIHS \cite{jojic2010structural} use photographic cameras, and thus, are useful to test and compare algorithms for visual lifelogging. Most of them are acquired using video (HTR cameras), making  the analysis of long periods of time difficult. Although nearly all of them show scenes of daily living and some of them record many continuous hours of video \cite{lu2013story,pirsiavash2012detecting,poleg2014temporal}, there is a strong need to create rich datasets with detailed annotations to ensure the robustness, applicability and usability of the algorithms for visual storytelling construction.

Following, we enumerate the available datasets (referenced by their main citation) for each of the relevant tasks applicable for analysing the main building blocks of lifelogging data:
\begin{itemize}
	\item Social interaction analysis: \cite{fathi2012social, alletto2014ego, narayan2014action}
	\item Object recognition/detection/discovery: \cite{bolanos2015ego-object, ren2009egocentric, fathi2011learning, pirsiavash2012detecting, behera2013egocentric, damen2014multi, bullock2015yale}
    \item Gaze prediction: \cite{fathi2012learning}
    \item Hand detection/segmentation: \cite{fathi2011learning, baraldi2014gesture, pirsiavash2012detecting, li2013pixel, egohands2015iccv, betancourt2015dynamic}
    \item Gesture recognition: \cite{baraldi2014gesture, bullock2015yale, cai2015scalable}
    \item Activity recognition: \cite{fathi2012learning, fathi2012social, poleg2014temporal, pirsiavash2012detecting, ryoo2013first, spriggs2009temporal, song2014activity, behera2013egocentric, iwashita2014first, kitani2011fast}
    \item Novelty or informative region detection: \cite{lu2013story, aghazadeh2011novelty}
\end{itemize}

This analysis reveals the lack of well-established and widely accepted datasets.
 

\subsection{Egocentric Vision Software}\label{sec:software}

The publication of the source code is crucial to guarantee the reproducibility of research results and to allow quantitative comparisons on different datasets.
To divulge available egocentric vision-related software, we present a list of the most relevant repositories, including source code for object recognition, object discovery, activity recognition, event segmentation, keyframe-based summarization and informative image detection in Table \ref{tab:software}. 

\vspace{-1em}
\section{Conclusions and Future Directions}\label{sec:conclusions}

This review summarized the state of the art of visual lifelogging analysis from a storytelling perspective, focusing on the progresses made so far in this context in the field of computer vision.
\textcolor{blue}{
In the first part of this survey we reviewed several techniques for acquiring, organizing, summarizing and browsing large collections of unstructured data. In the second part, we organized the available literature around the central questions necessary to address the storytelling problem: \textit{Was the user interacting  with somebody? How?}, \textit{Where} is he/she?,  \textit{When} did the event occur? and \textit{What} is the person wearing the camera doing?. For each research question we highlighted the weaknesses and strengths of available methods with respect to their applicability to the LTR domain. Additionally, we reviewed all the available datasets and source code.
}

Generally, from this review, we can draw some conclusions regarding the crucial points that must be followed in short-term research into egocentric vision. \textit{First}, there is a need to develop more algorithms suited to data acquired through photo cameras, in particular for social interaction detection and analysis, as well as for activity and context recognition.
 \textit{Second}, in view of the large number of datasets made publicly available in the last few years, it would be useful to foster cooperation within the lifelogging scientific community  to elaborate richer lifelogging datasets. 
By doing this, researchers could validate their algorithms and promote competition. 
 \textit{Third}, considering that visual storytelling has to preserve semantics, a promising direction is to continue leveraging semantic information for both egocentric data analysis and summarization. Given the wide variety of settings in which lifelogging cameras are being deployed, visual recognition could largely benefit from the use of ontologies. 
Moreover, this paper showed that the interest in 
analysis from the computer vision community over the last few years has increased considerably. In parallel, we witnessed a burst in the study and applicability of convolutional neural networks, suggesting that expectations for making progress in the coming years are growing fast.  
This progress should be accompanied by the creation of larger and more consolidated datasets that will compensate the enormous data demand of CNNs. In particular, 
\textcolor{blue}{research efforts should focus on the problems of 1) developing} more sophisticated transfer learning strategies able to reduce the need of large annotated datasets \textcolor{blue}{and 2)  exploiting temporal coherence of concepts that characterize visual lifelogs}. However, given the current limitations of CNNs in terms of computational cost and resources, the analysis would be limited to post-processing. 
Finally, a promising area of research that has not been explored for storytelling via ego-vision yet, is \emph{text description generation} from images. This problem, tackled for instance in \cite{yao2015describing,venugopalan2015sequence}, consists of rendering a visual to text translation of what is happening in the images. The development of these new kinds of multi-modal techniques could open up a new area, full of potential for egocentric storytelling, in which we could provide a human-like description of what happened in a precise scene or event. The application of these algorithms to the medical field, and more precisely to people with dementia, could help provide patients with a richer context to understand better what happened to them in a given situation.

\vspace{-1em}
\section*{Acknowledgments}

This work was partially funded by TIN2012-38187-C03-01, TIN2015-66951-C2-1-R and SGR 1219.
M. Dimiccoli is supported by a \textit{Beatriu de Pin\`os} grant (Marie-Curie COFUND action), and P. Radeva by an \textit{ICREA Academia} grant.

\ifCLASSOPTIONcaptionsoff
  \newpage
\fi



%




\bibliographystyle{plain}
\bibliography{mybibfile}

%





\begin{minipage}{\columnwidth}

\vspace{-4em}

\begin{IEEEbiography}
[{\includegraphics[trim=0 0.6cm 0 0.6cm, clip=true, width=1in,height=1in,clip,keepaspectratio]{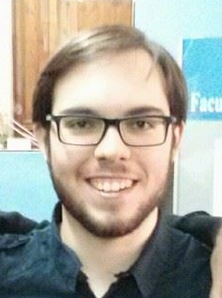}}]{Marc Bola\~nos} received his BSc in Computer Science from the Universitat de Barcelona (UB), Spain, in 2013. He received his interuniversity MSc in Artificial Intelligence from the Universitat Polit\`ecnica de Catalunya
in 2015 and is currently a PhD candidate at the UB. 
His research interests are in the area of deep neural networks and egocentric vision for health improvement.
\end{IEEEbiography}

\vspace{-4em}

\begin{IEEEbiography}
[{\includegraphics[trim=0 0.6cm 0 0.6cm, clip=true, width=1in,height=1in,clip,keepaspectratio]{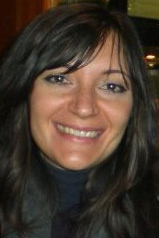}}]{Mariella Dimiccoli} is a Beatriu de Pin\`os fellow (Marie-Curie COFUND action) at the Computer Vision Center and associate professor at UB. She received a Computer Engineering degree (Cum Laude) from the Technical University of Bari and a PhD from the Universitat Polit\'ecnica de Catalunya (Cum Laude and European Honours). Her current research interests are the automatic analysis and organisation of lifelogging data acquired by a wearable camera and their use in health applications. 
\end{IEEEbiography}

\vspace{-4em}

\begin{IEEEbiography}
[{\includegraphics[width=1in,height=1in,clip,keepaspectratio]{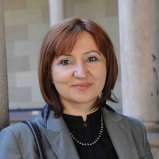}}]{Petia Radeva} is a tenure associate professor at UB. 
She is the head of the Consolidated Research Group "Computer Vision at the UB" and the head of MiLab at the Computer Vision Center (www.cvc.uab.es). Her research interests are in the development of learning-based approaches, especially in lifelogging applied to health.
She has published more than 200 international SCI journal papers and conference contributions. She is associate editor of the Journal of Visual Communication and Image Representation.
She received the ICREA Academia 2015 award from Catalonia given to the best 30 researchers of the year.
\end{IEEEbiography}

\vfill
\vfill
\vfill
\vfill

\end{minipage}

\break 
\pagebreak
\newpage




\vfill

\break
\pagebreak

\begin{@twocolumnfalse}

\begin{longtable}{p{3.5cm} p{8cm} p{3cm} R{1.5cm}}
	\caption{Summary of currently available public egocentric datasets.} \label{tab:datasets} \\
  	\hline 
    Name & Description & Type of Annotations & Camera \\ \hline \hline	
    
    Egocentric Dataset of the University of Barcelona (EDUB) \cite{bolanos2015ego-object} & With 4912 images acquired by the wearable camera Narrative; divided into 8 different days which capture daily life activities like shopping, eating, riding a bike, working, etc. It was acquired by 4 different subjects, 2 days each; and with 11,294 different object segmented instances from 21 different classes (TV, hand, person, car, sign, etc.). & Object labels and segmentations & Narrative \\ \hline
    
    All I Have Seen (AIHS) \cite{jojic2010structural} & Contains 19 days with a total of 45,612 images of 640 x 480 resolution, containing around 15 recurrent places/scenes appearing like home rooms, work office, work building, supermarkets, playgrounds, campus, biking trails, etc. & Not available & SenseCam \\ \hline
    
    Intel Egocentric Object Dataset \cite{ren2009egocentric} & Has 10 video sequences (100,000 frames) from 2 subjects manipulating 42 different types of everyday object instances. & Object labels and foreground and background segmentations & PointGrey \\ \hline
    
    GeorgiaTech Egocentric Activities (GTEA) \cite{fathi2011learning} & The videos captured by a cap-worn camera show 7 types of daily activities, such as making a sandwich/coffee/tea, each performed by 4 different subjects. Each activity video is labelled with the list of objects involved; each frame has left hand, right hand, and background segmentation marks & Objects list and hands and background segmentations & GoPro \\ \hline

	GTEA Gaze+ Dataset \cite{fathi2012learning} & With video and audio recordings of 7 meal-preparation activities such as making pizza/pasta/salad collected using eye-tracking glasses. Each activity was performed by 5 different subjects. Each frame has eye-gaze fixation data, and different activities such as “opening fridge” are annotated. & Gaze and actions performed & Tobii \\ \hline
    
    First-Person Social Interactions Dataset \cite{fathi2012social} & Day-long videos of 8 subjects spending their day at Disney World. The cameras are mounted on a cap worn by the subjects. Elan annotations containing the number of active participants in the scene, and the type of activity: walking, waiting, gathering, sitting, buying something, eating, etc. & Actions performed and social interactions at each time period & GoPro \\ \hline
    
    Huji EgoSeg Dataset \cite{poleg2014temporal}  & With 29 videos captured by an egocentric camera annotated in Elan format. The videos (some from YouTube and others recorded by Hebrew University of Jerusalem researchers) contain various daily activities. & Actions performed at each time period & GoPro \\ \hline
    
    UT Ego Dataset \cite{lu2013story} & Has 4 videos captured by a Looxcie wearable camera (head-mounted). Each video is about 3-5 hours long, captured in a natural, uncontrolled setting. The videos capture a variety of daily activities. & Important regions annotation & Looxcie \\ \hline
    
    Interactive Museum Dataset \cite{baraldi2014gesture} & A gesture recognition dataset taken from an egocentric perspective in a virtual museum environment. It has 5 different users who performed 7 hand gestures. & Hand gestures & No Information \\ \hline
    
    VINST - Visual Diaries \cite{aghazadeh2011novelty} & With 31 videos capturing the visual experience of a subject walking from a metro station to work. It consists of 7236 images in total. Each image is annotated with a location ID which covers 9 unique labels in total. Temporal segments corresponding to novel ego motions are annotated as well. & Location and "novel ego-motions" annotations per frame & No Information \\ \hline
    
    UCI Activities of Daily Living Dataset (ADL) \cite{pirsiavash2012detecting} & Has 1 million frames of dozens of people performing 18 daily indoor activities such as brushing their teeth, washing dishes, or watching television, each performed by 20 different subjects. It includes annotations of 42 object classes. & Activities, object bounding boxes and classes, hand positions and interaction events & GoPro \\ \hline
    
    EGO-HPE \cite{alletto2014ego} & A set of egocentric videos with different subjects for head pose estimation. Each video is annotated at the frame level for five yaw angle orientations (-75º, -45º, 0º, 45º, 75º) with respect to the subject wearing the camera. & Face orientation & Vuzix Smart Glass \\ \hline
    
    EGO-GROUP \cite{alletto2014ego}& A social group detector dataset for egocentric vision, which consists of 10 videos collected in different situations: a laboratory, a coffee break, a conference room and an outdoor scenario. & People group composition & Vuzix Smart Glass \\ \hline
    
    JPL First-Person Interaction Dataset \cite{ryoo2013first} & Human activity videos taken from a first-person viewpoint. The dataset specifically aims to provide first-person videos of interaction-level activities, recording how things look from the perspective of a person/robot participating in physical interactions. & Actions performed in each time period & GoPro \\ \hline
    
    NUS First-person Interaction Dataset \cite{narayan2014action} & Dataset for interaction recognition with 8 interactions in 2 perspectives (first-person and third-person) resulting in 16 classes in total. The dataset will be made publicly available at a later date. It contains 2 human-human interactions, 2 human-object-human interactions and 4 human-object interaction classes. It contains 260 videos with at least 15 samples in each class. & Interaction type & GoPro \\ \hline
    
    CMU Multi-Modal Activity Database (CMU-MMAC) \cite{spriggs2009temporal} & Multimodal dataset of 18 subjects cooking 5 different recipes (brownies, pizza, etc.); also contains audio, body motion capture, and IMU data. & Frame-level action & No Information \\ \hline
    
	CMU EDSH (hands under varying illuminations) \cite{li2013pixel} & Dataset of over 600 hand images taken under various illumination conditions and different backgrounds. Each image is segmented at the pixel level. & Hand segmentation & GoPro  \\ \hline
    
    EgoHands Dataset \cite{egohands2015iccv} & Contains 48 Google Glass videos of complex, first-person interactions between two people. The main intention of this dataset is to enable better, data-driven approaches to understand hands in first-person computer vision. & Hand segmentation & Google Glass \\ \hline
    
    Unige-Hands Dataset \cite{betancourt2015dynamic} & Videos recorded in 5 different locations (office, street, bench, kitchen and coffee bar) intended for hand detection. & Hand/No Hand label per frame & GoPro \\ \hline
    
    Yale Human Grasp Dataset \cite{bullock2015yale} & Dataset with 27.7 hours of tagged video recorded by two housekeepers and two machinists during their regular work activities. It includes the tagged grasp type with its time information, objects manipulated and parameters of the performed task. & Grasp tagging, and interval and object labels & RageCams \\ \hline
    
    UT Grasp Data Set \cite{cai2015scalable} & Dataset under controlled environment performed by four different subjects. They were asked to grasp a set of objects placed on a desktop with specific types of grasps. The most common subset of 17 grasp types from Feix's Taxonomy \cite{feix2009comprehensive} 
were selected to perform these everyday activities. & Hand grasp type and start/end frame number & GoPro \\ \hline
    
    Life-logging EgoceNtric Activities (LENA) \cite{song2014activity} & Egocentric video database containing 13 categories of activities relevant to lifelogging applications performed by 10 different subjects. Each subject recorded 2 clips for one activity (20 clips per activity). Each clip has a duration of 30 seconds. & Activities performed. & Google Glass \\ \hline
    
    COGNITO \cite{behera2013egocentric} & Non-periodic manipulative tasks in an industrial context. All the video sequences were captured with on-body sensors consisting of IMUs, a backpack-mounted RGB-D camera for top-view and a chest-mounted fish-eye camera for the front view of the workbench. & Activity labels and objects and wrist tracklets & RGB-D and others \\ \hline
    
    Michigan-Milan Indoor Dataset \cite{furlan2013free} & With 10 video sequences collected with common smartphones in a variety of environments, including offices, corridors and large rooms, where the observer moves freely (6 DoF) around the scene. & Image segmentations with the labels "ceiling", "floor" or "wall" & Smartphone \\ \hline
    
    Bristol Egocentric Object Interactions Dataset \cite{damen2014multi} &  Dataset captured with wearable gaze tracker software containing various pre-defined actions of daily living in different indoor locations (kitchen, workspace, gym, laser printer, corridor and weight-lifting machine). The videos in each sequence are recorded by 3-5 different users. & 3D maps and 3D objects GT & ASL Mobile Eye XG \\ \hline
    
    DogCentric Activity Dataset \cite{iwashita2014first} &  DogCentric Activity Dataset is composed of dog activity videos taken from a first-person animal viewpoint. The dataset contains 10 different types of activities, including activities performed by the dog itself, interactions between people and the dog, and activities performed by people or cars. The videos are in 320x240 image resolution, 48 frames per second. & Activity performed & GoPro \\ \hline
    
    UEC EgoAction Dataset \cite{kitani2011fast} & A set of videos (acquired by the researchers or public from YouTube) recording different sports (skiing, mountain biking, etc.). Each video is several minutes long and contains a wide set of actions performed by the user. & Activities performed & GoPro \\ \hline

\end{longtable}

\end{@twocolumnfalse}

\pagebreak
\newpage
\mbox{}
\vspace{10em}
\vfill
\break

\begin{@twocolumnfalse}

\begin{longtable}{p{5cm} p{12cm}}
	\caption{List of the most relevant public software related to egocentric vision.} \label{tab:software} \\
  	\hline 
    
    Alireza Fathi\textquotesingle s Egocentric Vision Toolbox \cite{fathi2012learning,fathi2011learning,ren2010figure} & Toolbox including functions for applying different data processing to egocentric videos, including motion estimation, image segmentation, object classification and action classification among others. \\
    OpenCV and CUDA & \url{http://ai.stanford.edu/~alireza/GTEA_Gaze_Website/Code/index.html} \\
    \hline
    
    Ego-Object Discovery \cite{bolanos2015ego-object,bolanos2015object} & Object Discovery Algorithm on Egocentric Images. Semi-supervised algorithm that uses initial object proposal generation, a CNN-based feature representation, false positive filtering, and an interactive object discovery with Refill strategy. \\
     Matlab and Caffe & \url{https://github.com/MarcBS/Ego-Object_Discovery} \\
    \hline
    
    Detecting Activities of Daily Living in First-person Camera Views \cite{pirsiavash2012detecting} & Train and test code for the problem of detecting activities of daily living (ADL). It applies novel representations including temporal pyramids to approximate temporal correspondences, and composite object models that exploit the differences between the objects when being interacted with. \\
   Matlab & \url{http://people.csail.mit.edu/hpirsiav/codes/ADLdataset/adl.html} \\
    \hline
    
    Temporal Pooling of CNN Vectors \cite{ryoo2014pooled} & It includes the pooled time series (PoT) representation framework as well as basic per-frame descriptor extractions including a histogram of optical flows (HOF) and histogram of oriented gradients (HOG). \\
    Java and OpenCV [exec. only] & \url{https://github.com/mryoo/pooled_time_series/} \\
    \hline
    
    Temporal Segmentation of Egocentric Videos \cite{poleg2014temporal} & Software for segmentation and event classification of egocentric HTR videos. It applies a hierarchical classification using cumulative displacement curves. \\
    Matlab and C++ & \url{http://www.vision.huji.ac.il/egoseg/} \\
    \hline
    
	Doherty Wearable Camera Browser \cite{Doherty:2008} & Application for data segmentation annotation and browsing. It supports analysis of images from the following photographic cameras: Vicon Autographer, Revue, or SenseCam. \\
    ~[exec. only] & \url{http://sensecambrowser.codeplex.com/} \\
    \hline
    
    R-Clustering for Event Segmentation \cite{talavera2015r-clustering} & Segmentation of events in egocentric lifelogging photo streams. It uses convolutional neural network features and an energy minimisation (Graph-Cut) technique to segment photo sequences. \\
    Matlab and Caffe & \url{https://github.com/MarcBS/SR-Clustering} \\
    \hline
    
    Motion-Based Egocentric Segmentation \cite{bolanos2014video} & It applies a robust SIFT-Flow motion estimation suitable for photo sequences to perform photo stream segmentation in motion-related events. \\
    Matlab & \url{https://github.com/MarcBS/Motion_Video_Segmentation} \\
    \hline
    
    Egocentric Vision Keyframe Summarization \cite{bolanos2015visual} & The code extracts a visual summary of a set of egocentric images captured by a photo camera. The result is a collage with one image summarizing every event in the image set. It uses a frame representation by means of a convolutional neural network followed by an event segmentation based on agglomerative clustering and keyframe selection based on Random Walk. \\
   Matlab and Caffe & \url{https://github.com/MarcBS/Egocentric-Visual-Keyframes-Summary} \\
    \hline

	Egocentric Snap Points Detection \cite{xiong2014detecting} & Automatic prediction of “snap points” in unedited egocentric video that is, those frames that look as if they could be photos taken intentionally. It makes use of a generative model for snap points that rely on a photo prior to intentional (conventional) images together with domain-adapted features. \\
    Matlab and C & \url{https://github.com/bxiong1202/snap-points} \\
    \hline

\end{longtable}

\end{@twocolumnfalse}

\pagebreak
\newpage
\mbox{}
\vspace{10em}
\vfill
\break

\pagebreak
\vspace{10em}
\vfill
\break

\end{document}